\pdfoutput=1

\documentclass[11pt]{article}

\usepackage[preprint]{latex/acl}
\usepackage{makecell}
\usepackage{xcolor}
\usepackage{graphicx}
\usepackage{subfigure}
\usepackage{stfloats}
\usepackage{color}
\usepackage{pifont}
\usepackage{balance}
 
\definecolor{midnightgreen}{rgb}{0.0, 0.29, 0.33}

\usepackage{times}
\usepackage{latexsym}
\usepackage{amsmath}
\usepackage[T1]{fontenc}

\usepackage[utf8]{inputenc}
\usepackage{amssymb}
\usepackage{microtype}

\usepackage{inconsolata}
\usepackage{multirow}
\usepackage{graphicx}
\title{Say More with Less: Understanding Prompt Learning Behaviors through Gist Compression}

\author{Xinze Li$^{1}$, Zhenghao Liu$^{1}$\thanks{ \ \ indicates corresponding author.}, Chenyan Xiong$^{2}$, Shi Yu$^{3}$, Yukun Yan$^{3}$, Shuo Wang$^{3}$ and Ge Yu$^{1}$ \\ 
$^1$Department of Computer Science and Technology, Northeastern University, China \\
$^2$Language Technologies Institute, Carnegie Mellon University, United States\\
$^3$Department of Computer Science and Technology, Institute for AI, Tsinghua University, China \\
Beijing National Research Center for Information Science and Technology, China \\
}

\begin{document}
\maketitle
\begin{abstract}
Large language models (LLMs) require lengthy prompts as the input context to produce output aligned with user intentions, a process that incurs extra costs during inference. In this paper, we propose the \textbf{Gist} \textbf{CO}nditioned de\textbf{CO}ding (Gist-COCO) model, introducing a novel method for compressing prompts which also can assist the prompt interpretation and engineering. Gist-COCO employs an encoder-decoder based language model and then incorporates an additional encoder as a plugin module to compress prompts with inputs using gist tokens. It finetunes the compression plugin module and uses the representations of gist tokens to emulate the raw prompts in the vanilla language model. By verbalizing the representations of gist tokens into gist prompts, the compression ability of Gist-COCO can be generalized to different LLMs with high compression rates. Our experiments demonstrate that Gist-COCO outperforms previous prompt compression models in both passage and instruction compression tasks. Further analysis on gist verbalization results suggests that our gist prompts serve different functions in aiding language models. They may directly provide potential answers, generate the chain-of-thought, or simply repeat the inputs. All data and codes are available at \url{https://github.com/OpenMatch/Gist-COCO}.



\end{abstract}
\section{Introduction}

Large Language Models (LLMs), such as GPT-4~\cite{achiam2023gpt} and LLaMA~\cite{touvron2023llama}, have demonstrated their emergent capacity in handling various NLP tasks~\cite{zhao2023survey,wei2022emergent}. To align user intentions with LLMs, existing work pays increasing attention to prompt engineering. They attempt to optimize prompts using LLMs themselves or manually craft prompts with meticulous care~\cite{zhou2022large,cheng2023black,ye2023investigating}. Nevertheless, the challenge persists in deciphering user intentions from natural language by LLMs and providing explainable insights for prompt engineering.

\begin{figure}[t]
\centering
 \includegraphics[width=\linewidth]{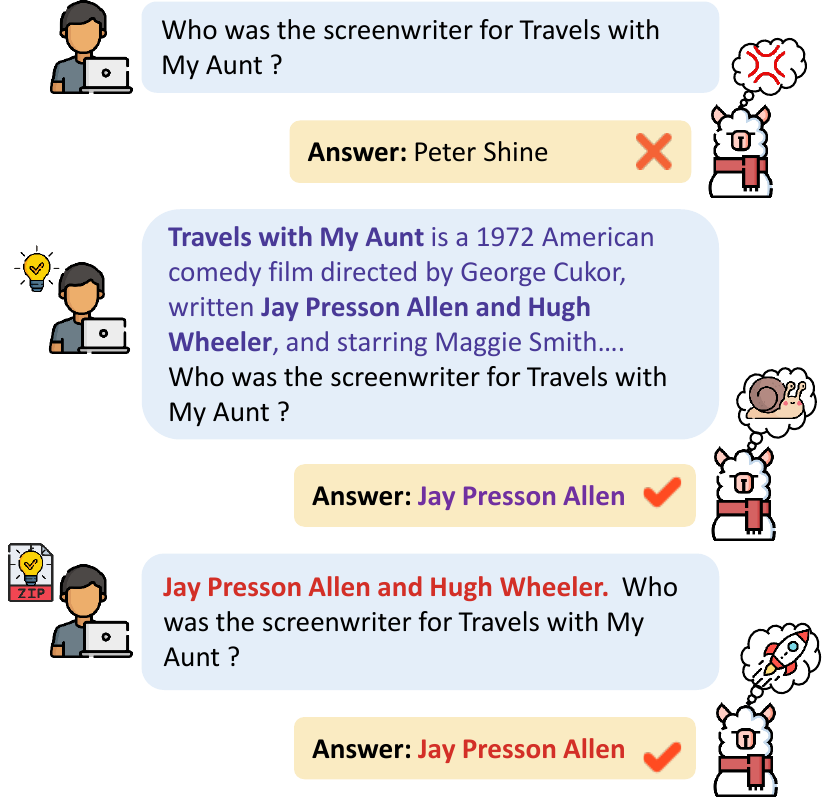}\label{fig:embed:CL}
\caption{The Motivation of Our Gist Conditioned Decoding (Gist-COCO) Model. The user respectively utilizes prompts\includegraphics[width=1.2em]{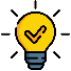} and compressed prompts\includegraphics[width=1.2em]{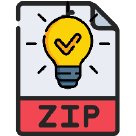} to guide the generation of LLMs.}
\label{fig:introduction}
\end{figure}

As shown in Figure~\ref{fig:introduction}, users typically collect or compose detailed prompts to assist LLMs in generating answers, making them more tailored and precise. 
However, with each user query, LLMs must iteratively encode these prompts and compute their self-attention~\cite{vaswani2017attention}, leading to increased computational time and memory usage~\cite{mu2023learning}. Reducing the length of prompts is a potent strategy to optimize these prompts. Existing work utilizes the theory of self-information~\cite{shannon1948mathematical} to explain prompts and reduce them by filtering the contexts with low self-information in the prompts~\cite{li2023unlocking}. \citet{mu2023learning} further compress task instructions by utilizing gist tokens and employing the resulting gist embeddings for instruction representation. Nevertheless, achieving interpretability and refinement in prompt compression, which is crucial for prompt engineering and understanding LLMs' behavior, remains challenging yet.


To alleviate the problem, this paper introduces the \textbf{Gist} \textbf{CO}nditioned de\textbf{CO}ding (Gist-COCO) model, which targets on compressing prompts and generalizing compression to different LLMs. Our Gist-COCO model is inspired by information theory~\cite{grunwald2007minimum} and built upon an encoder-decoder based language model, such as FlanT5~\cite{chung2022scaling}. It employs an extra encoder model as a compression plugin module to compress prompts with inputs using a set of shorter gist tokens whose representations are utilized to replace the raw prompts of inputs. Specifically, these gist representations are contacted as prefixes with the input representations encoded by the vanilla encoder and fed into the vanilla decoder. Gist-COCO only finetunes the compression model to generate more effective gist representations, aiding the vanilla FlanT5 model in adhering closely to the raw prompts for the generation. Additionally, our Gist-COCO model incorporates a task disentangled gist modeling method to effectively compress various types of prompts, such as passages and instructions. 

To generalize the compression capabilities of Gist-COCO across different LLMs, we propose the gist verbalization method, which can verbalize gist representations into some shorter gist prompts using the language model. By preprocessing the prompts with inputs using the compression module, the gist prompts refine the essential information from the raw prompts based on the inputs. Instead of using annotated summarization data to learn prompt compression~\cite{vig2021exploring,xu2023recomp}, compression models, such as Gist~\cite{mu2023learning} and Gist-COCO, compress prompts using gist tokens and optimize these gist representations using vanilla prompts from training data. 
Additionally, unlike baseline models~\cite{mu2023learning,chevalier2023adapting}, our Gist-COCO model freezes the parameters of language models and only finetunes the encoder model for compression, which can generalize its compression ability.


Our experiments demonstrate the effectiveness of the Gist-COCO model, surpassing prior prompt compression models in both passage and instruction compression tasks. Leveraging our gist verbalization method, Gist-COCO broadens its advantages to different language models, achieving an exceptionally high compression rate. Besides, the results of gist verbalization show that gist prompts serve diverse roles in assisting language models to comprehend human instructions, such as encompassing the formation of answers, generating the thought, and copying parts of contents from inputs or instructions for reinforcement.

\section{Related Work}
Large Language Models (LLMs)~\cite{brown2020language}, typically finetune through instruction learning methods~\cite{chung2022scaling,openai2022chatgpt,taori2023stanford,chiang2023vicuna}, such as instruction tuning or Reinforcement Learning with Human Feedback (RLHF)~\cite{ouyang2022training}, can enhance their ability to adhere to instructions or align with human preferences. Besides, finetuning language models on diverse instruction-response pairs enables language models to exhibit cross-task generalization~\cite{wei2021finetuned,sanh2021multitask}. In this case, existing work focuses more on generating more instruction data~\cite{wang2022self,wan2023explore,mishra2021cross} or the task sensitive tasks~\cite{kung2023active} for supervised finetuning (SFT) LLMs.

To enhance the effectiveness of LLMs in downstream tasks, researchers are increasingly emphasizing prompt engineering~\cite{liu2023pre}. The prompts can serve as instructions to elucidate user intentions~\cite{zhou2022large} or provide the contextual knowledge to aid in the generation process~\cite{izacard2022few,ram2023context,tonmoy2024comprehensive,shi2023replug}. However, the prompts have demonstrated that they potentially exert a substantial influence on the LLMs' outputs~\cite{lu2022fantastically} and necessitate meticulous designs~\cite{chen2023unleashing,kaddour2023challenges}. 

To make prompts better guide the generation of LLMs, existing work focuses more on conducting more effective prompts in different ways. \citet{zhou2022large} use LLMs for automatic instruction generation and selection. \citet{cheng2023black} propose the Black-box Prompt Optimization (BPO) method, which optimizes the prompts to bridge the gap between humans and LLMs. \citet{ye2023investigating} further add the task-agnostic prefix to enhance the instruction. Nevertheless, it remains unclear which aspects of these provided prompts are favored by LLMs for comprehending human intentions.

Studying the characteristics of prompts in prompting LLMs has garnered much attention from researchers~\cite{min2022rethinking,beurer2023prompting}. The researchers use the Turking Test~\cite{efrat2020turking} and the negated prompts~\cite{jang2023can} to analyze the instruction understanding and following ability of LLMs. Instead of evaluating such an ability of LLMs, inspired by the minimum description length (MDL) principle~\cite{grunwald2007minimum}, we focus more on interpreting the role of prompts from a compression view.
Some existing work has shown effectiveness in prompt compression, \textit{e.g.} distilling the prompt understandings from teacher models to student models~\cite{snell2022learning}, compressing the prompts using a set of gist tokens~\cite{mu2023learning,ge2023context,chevalier2023adapting} and generating some brief summaries~\cite{vig2021exploring,xu2023recomp}. Based on these works, we aim to compress prompts as gist representations according to the need of language models and further verbalize them into gist prompts to interpret and understand the role of prompts.

\section{Methodology}
In this section, we first introduce prompt compression through the information theory (Sec.~\ref{model:pre}). We then describe our \textbf{Gist} \textbf{CO}nditioned de\textbf{CO}ding (Gist-COCO) model (Sec.~\ref{model:gist}). Finally, we show how to generalize the compression ability to different tasks and language models (Sec.~\ref{model:verbalization}).

\subsection{Preliminary of Prompt Compression}\label{model:pre}
Given an input $x$, existing work usually uses lengthy task instructions~\cite{wang2022self,chung2022scaling} or retrieved passages~\cite{yu2023augmentation,shi2023replug} as prompts, denoted as $c$, to aid LLMs for the generation.
To reduce inference cost, Gist-COCO compresses the raw long prompt $c$ into a few gist representations $h^c = \{h^c_1, ..., h^c_N\}$, serving as condensed context for LLM inference. 

Inspired by the compression viewpoint of the minimum description length (MDL) principle~\cite{grunwald2007minimum} in information theory~\cite{wu2022self}, a good model should be able to represent the data with shorter descriptions and also generalize well to unseen data~\cite{wu2022self}.
The MDL principle indicates that the best compression model $M^*$ can make the correct prediction $y^*$ based on a shorter codelength:
\begin{equation}\label{eq:mdl}
\small
M^*_\theta = \arg\min_{\theta} {L(M_\theta)+L(y^*|M_\theta(c,x))},
\end{equation}
where $L(M_\theta)$ is the codelength (model complexity) required by the model and $L(y^*|M_\theta(c,x))$ is the codelength to construct the correct prediction based on the compression result. 
The compression model $M_\theta$ encodes the prompt $c$ into a fixed number of hidden states $h^c$, given $c$ with the input $x$:
\begin{equation}
\small
    h^c = \{h^c_1, ..., h^c_N\} \leftarrow
    M_\theta(c,x) .
\end{equation}
As we fix $|h_c|=N$, the term $L(M_\theta)$ becomes constant in Eq.~\ref{eq:mdl} and our goal is to minimize $L(y^*|M_\theta(c,x))$.
In the next section, we introduce $M_\theta$ as well as its training and inference.
\begin{figure}[t]
\centering
 \includegraphics[width=\linewidth]{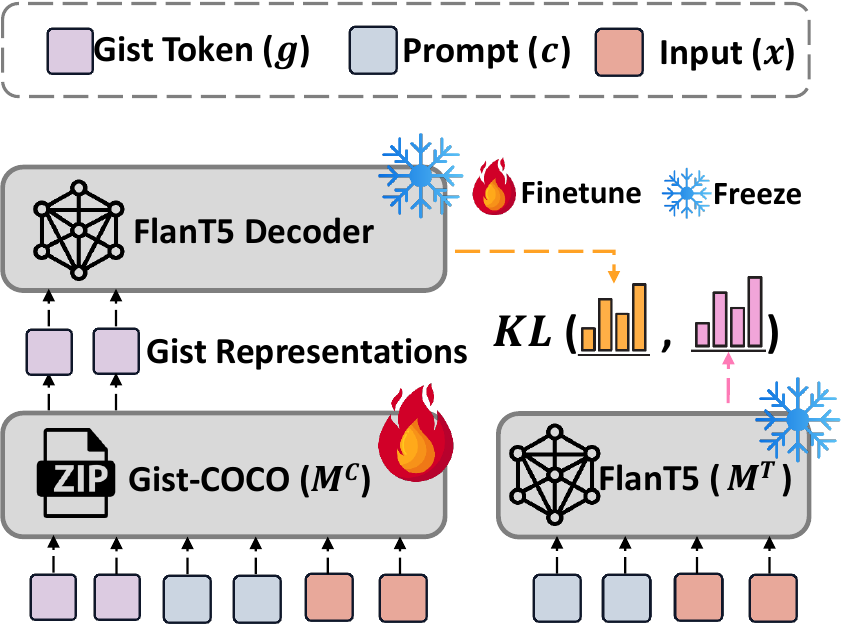}
\caption{Training of Gist-COCO. 
Gist-COCO is trained to emulate the output distribution based on uncompressed inputs by producing gist representations.
 }
\label{fig:model}
\end{figure}

\subsection{Prompt Compression via Gist Conditioned Decoding}\label{model:gist}

Given the prompt $c$ and input $x$, Gist-COCO is trained to minimize Eq.~\ref{eq:mdl} to produce the optimal gist representations $h^c = \{h^c_1, ..., h^c_N\}$ for the prompt $c$. 
As shown in Figure~\ref{fig:model}, we propose to leverage the soft labels from a vanilla language model $M^T$ with raw prompts to estimate the codelength with the help of Kullback-Leibler (KL) divergence between the uncompressed distribution and the compressed one:
\begin{equation}
    \small
L(y^*|M_\theta(c,x)) \approx \text{KL}(P(y^*|h^c,x) || Q(y^*|c,x)),
\end{equation}
where $P(y^*|h^c,x)$ is the generation probability given the gist representations calculated from FlanT5-Decoder, and $Q(y^*|c,x)$ is the prior from the model given raw prompts, calculated from $M^T$ (FlanT5):
\begin{equation}
\small
\begin{aligned}
    P(y^*|h^c,x) &= 
    \text{T5-Decoder} (h^c), \\
    Q(y^*|c,x) &= 
    \text{$M^T$}(c;x),
\end{aligned}
\end{equation}
where $;$ denotes concatenation. The parameters of $M^T$ are frozen during training.
$h_c$ is encoded by the compression model $M^C_\theta$, which is initialized with the same parameters as the model $M^T$:
\begin{equation}
    \small
    {h}^c \leftarrow M^C_\theta (c,x)=\text{T5-Encoder}(\{g_1, ..., g_N\};c;x),\label{eq:compress_hidden}
\end{equation}
where $\{g_1, ..., g_N\}$ are the gist tokens to compress the prompt $c$, whose weights are initialized from the special tokens of the FlanT5 model. ${h}^c$ are the encoded representations of $\{g_1, ..., g_N\}$ using $M^C$.


During inference, following Eq.~\ref{eq:compress_hidden}, we use the trained compression model $M_\theta^C$ to compress the prompt to obtain gist representations $h^c$, and feed them with the encoded input $x$ into the decoder to obtain the output:
\begin{equation}
\small
    y = \text{T5-Decoder} (h^c;\text{T5-Encoder}(x)).
\end{equation}

\subsection{Compression Generalization for Different Prompts and Language Models}\label{model:verbalization}
In this subsection, we generalize Gist-COCO to different tasks and language models by task disentangled gist modeling and prompt verbalization.

\textbf{Task Disentangled Gist Modeling.} We compress two types of prompts during modeling, including retrieved passages~\cite{guu2020retrieval} and instructions~\cite{mu2023learning}, which are typically used in existing language models.

For instruction compression, we regard the task instruction as the prompt $c$ and then use $N$ instruction gist tokens $\{g^{i}_1, ..., g^{i}_N\}$ for compression:
\begin{equation}\label{eq:compress_hidden_1}
\small
    h^c \leftarrow M^C (\{g^{i}_1, ..., g^{i}_N\};c;x),
\end{equation}
where $h^c = h^c(g^i)$. $h^c(g^i)$ represents the set of encoded representations of $\{g^{i}_1, ..., g^{i}_N\}$. 
In the retrieval-augmented generation (RAG) models, we regard the concatenation of retrieved passages and task instructions as the prompt $c$. Then we use both $N$ passage gist tokens $\{g^{p}_1, ..., g^{p}_N\}$ and $N$ instruction gist tokens $\{g^{i}_1, ..., g^{i}_N\}$ for compression:
\begin{equation}\label{eq:compress_hidden_2}
\small
    h^c \leftarrow M^C (\{g^{p}_1, ..., g^{p}_N\};\{g^{i}_1, ..., g^{i}_N\};c;x),
\end{equation}
where $h^c = \{h^c(g^p); h^c(g^i)\}$. $h^c(g^p)$ and $h^c(g^i)$ are the compressed representations of the passage gist tokens $g^p$ and the instruction gist tokens $g^i$.

\textbf{Gist Verbalization.} To generalize the advantages of our Gist-COCO model to decoder-based language models, we use the vanilla FlanT5 decoder to decode the compressed hidden states $h^c$ to get the gist prompts $v = \{v_1,...,v_k\}$:
\begin{equation}\label{eq:verbalization}
\small
v= \text{T5-Decoder}(h^c).
\end{equation}
We can assess the compression effectiveness of our Gist-COCO model by replacing the prompt $c$ with the shorter gist prompts $v$ when utilizing decoder-based language models. Besides, we can further observe and understand the effectiveness of prompt learning by analyzing the gist prompts $v$.

\section{Experimental Methodology}
This section describes the datasets, evaluation metrics, baselines, and implementation details.

\textbf{Dataset.} In our experiments, we use different datasets to build the training and evaluation benchmarks. All data statistics are shown in Table~\ref{tab:dataset}.

\begin{table}
\begin{center}
\small
\begin{tabular}{l|l|l|r}
\hline
                           \textbf{Split}    &\textbf{Dataset}  &\textbf{Setting}  &\textbf{Total}  \\ \cline{1-4}
\multirow{2}{*}{Training} & \multirow{2}{*}{NVI2}   &   Instruction   & 94,481  \\
                       &                         &Passage      & 92,607   \\\cline{1-4}
\multirow{7}{*}{Evaluation}  & PopQA &  -        & 14,267    \\\cline{2-4}
                       & \multirow{3}{*}{KILT}   & NQ         & 2,837    \\
                       &                         & TrivialQA & 5,359    \\
                       &                         & HotpotQA   &  5,600      \\\cline{2-4}
                       & \multirow{3}{*}{Alpaca+} & Seen       & 1,000       \\
                       &                         & Unseen     & 1,000      \\
                       &                         & Human      & 252    \\
\hline
\end{tabular}

\caption{Data Statistics.}
\label{tab:dataset}
\end{center}
\end{table}

\textit{Training.} During training Gist-COCO model, we use Natural Instruction v2 (NVI2)~\cite{wang2022benchmarking} dataset to build the training set for compression. The training dataset consists of instruction compression and retrieved passage compression tasks. For instruction compression, we filter out the non-English tasks and reserve 1,053 tasks. We randomly sample up to a maximum of 90 instances from each task, resulting in a total of 94,481 pieces of data. For the retrieved passage compression, we selected 30 tasks from NVI2 dataset, amounting to a total of 92,607 pieces of data. These selected tasks usually require external knowledge and we use T5-ANCE~\cite{yu2023openmatch,yu2023augmentation} to retrieve passages from MS MARCO~\cite{nguyen2016ms} for augmenting the language model.

\textit{Evaluation.} During evaluation, we use different datasets to estimate the effectiveness of retrieved passage compression and instruction compression.


Following previous work~\cite{mu2023learning}, we use Alpaca+ dataset~\cite{mu2023learning} to evaluate the instruction compression effectiveness of Gist-COCO. The Alpaca+ dataset is a large instruction finetuning dataset, which combines both Self-Instruct~\cite{wang2022self} and Stanford Alpaca~\cite{taori2023stanford} datasets. To evaluate the effectiveness of retrieved passage compression, we use PopQA~\cite{mallen2023not} as well as NQ~\cite{kwiatkowski2019natural}, TrivialQA~\cite{joshi2017triviaqa} and HotpotQA~\cite{yang2018hotpotqa} from KILT~\cite{petroni2020kilt} for evaluation, where we use the dev set for all tasks from KILT. The KILT-Wikipedia~\cite{petroni2020kilt} is regarded as the knowledge base for seeking knowledge. Then we use T5-ANCE~\cite{yu2023openmatch,yu2023augmentation} to retrieve passages from it for augmentation.


\begin{table*}[t]

\centering
\small

\begin{tabular}{l|l|c|ccc|ccc}
\hline
\multirow{3}{*}{\textbf{LLM}} & \multicolumn{1}{l|}{\multirow{3}{*}{\textbf{Method}}}  & \multicolumn{4}{c|}{\textbf{Passage Compression}} & \multicolumn{3}{c}{\textbf{Instruction Compression}} \\  \cline{3-9}   &   & \multirow{2}{*}{\textbf{PopQA}}  & \multicolumn{3}{c|}{\textbf{KILT}}  & \multicolumn{3}{c}{\textbf{Alpaca+}} \\\cline{4-9}  

                                            &                        &            &  NQ        & TrivialQA        & HotpotQA   &  Seen        & Unseen   &Human                   \\ \hline
\multirow{6}{*}{FlanT5-base}          & No Prompt                                                   & 8.8         &  4.4       &  9.2        & 12.1    &  20.3        &  22.0     &  9.7                    \\
                                       & AutoCompressor ~\shortcite{chevalier2023adapting}                     &  8.3        &   4.8     &   9.4      &   12.2  &20.3 & 22.7 & 7.8          \\
                                       & Gist ~\shortcite{mu2023learning}                                &    8.4      &  4.6        & 8.9    & 12.1     &18.3  &18.9 &  8.7        \\
                                & Gist (Ours)~\shortcite{mu2023learning}    &9.4  &5.7 &11.4 &11.6&23.1&27.5&\textbf{14.1}\\
                                            & Gist-COCO                                &\textbf{31.0}        & \textbf{22.9}        & \textbf{50.9}       &\textbf{17.2}     &\textbf{23.6}	&\textbf{29.0}	&12.1

                                            \\ \cline{2-9}

                                        & Full Prompt                           &  43.9        &  30.0      &   61.9     & 23.2   &  23.9      &   29.8     &  15.2     \\ \hline

\multirow{6}{*}{FlanT5-large}          & No Prompt                                                   &  7.3        & 8.3       &  19.0        &     14.6  & 19.2     &  18.4       &  10.3              \\
                                       & AutoCompressor ~\shortcite{chevalier2023adapting}                      &   5.8       &  8.4      &    19.1     &  14.7  &16.6   & 12.2 &  6.6           \\
                                       & Gist   ~\shortcite{mu2023learning}                            &    9.1     &  8.2          &  18.9      & 14.6    & 21.4 &19.0 &  10.7        \\

                                        & Gist (Ours) ~\shortcite{mu2023learning}  &11.6&8.4&19.1&13.0&24.3&29.4 &\textbf{15.7}\\
                                    
                                        & Gist-COCO  &\textbf{32.0} &\textbf{27.0} &\textbf{57.3} &\textbf{20.6} 
                                       & \textbf{25.7}	&\textbf{30.1}	&14.0

                                        \\ \cline{2-9}
                                        & Full Prompt                          &  46.0       &   34.4     &  67.1   & 27.5    &  26.7      &  32.3      &   18.8                 \\ \hline

\end{tabular}

\caption{Overall Performance of Different Prompt Compression Methods.}
\label{tab:overall}
\end{table*}

\textbf{Baselines.} 
In our experiment, we compare our Gist-COCO model with several baselines.

Two embedding based compression models are compared in our experiments, including AutoCompressor~\cite{chevalier2023adapting} and Gist~\cite{mu2023learning}. We directly use the AutoCompressor and Gist models to compress the prompts as representations and then train a linear layer to adapt the compressed representations to the FlanT5 model. AutoCompressor is an unsupervised model, which compresses long contexts into a set of summary vectors to facilitate different generation tasks. Different from AutoCompressor, Gist~\cite{mu2023learning} is a supervised method, which finetunes the language models on the Alpaca+ instruction dataset and teaches the model to compress the instructions through the attention mask. 

Besides, we also reimplement the Gist model, denoted as Gist (Ours), maintaining identical model architecture with~\citet{mu2023learning}. We finetune this model using the same training dataset employed for our Gist-COCO model. Furthermore, we utilize the SEGENC model~\cite{vig2021exploring} as a baseline, which finetunes BART~\cite{lewis2020bart} model using the query-focused summarization dataset.

\textbf{Evaluation Metrics.}
Following previous work~\cite{mu2023learning}, we used the ROUGE-L metric to evaluate the performance of different models on instruction compression tasks. For the passage compression tasks, we use accuracy as an evaluation metric, which is similar to~\citet{yu2023augmentation}. We conduct string matching between the generated answer and the golden answer.

\textbf{Experimental Details.}
This part describes the experiment details of Gist-COCO model. 

We initialize Gist-COCO model with FlanT5-base and FlanT5-large checkpoints from Hugginface Transformers~\cite{wolf2019huggingface}. During training, we use the top-1 ranked passage from retrieval as the prompt to enhance the generation results for these passage compression tasks. In our experiments, we set the learning rate as 1e-4 and the training epoch as 8. During inference, we use the top-5 ranked passages from retrieval as the prompt for all passage compression tasks.


\section{Evaluation Results}
In this section, we first evaluate the performance of Gist-COCO on passage and instruction compression tasks. Subsequently, we conduct ablation studies and further analyze the characteristics of learned gist representations. Finally, the case studies are presented.

\begin{table*}[t]
\small
\centering
\resizebox{\linewidth}{!}{
\begin{tabular}{l|l|c|ccc|c|ccc|c}
\hline
\multirow{3}{*}{\textbf{LLM}} & \multirow{3}{*}{\textbf{Method}}& \multicolumn{5}{c|}{\textbf{Passage Compression}}& \multicolumn{4}{c}{\textbf{Instruction Compression}} \\  \cline{3-11}

& & \multirow{2}{*}{\textbf{PopQA}}  & \multicolumn{3}{c|}{\textbf{KILT}}& \multirow{2}{*}{\textbf{Ratio}} & \multicolumn{3}{c|}{\textbf{Alpaca+}} & \multirow{2}{*}{\textbf{Ratio}} \\ \cline{4-6}\cline{8-10}
                                            &                        &             &  NQ        & TrivialQA        & HotpotQA  &   & Seen & Unseen   &Human  &               \\ \hline

\multirow{4}{*}{Llama-7b}          & No Prompt                                                   & 22.8         &  20.5       &  61.7        &    18.2  & - & 21.7  	&  21.1 & 4.6 & -          \\

                                       & SEGENC~\shortcite{vig2021exploring}                             &  25.9        &  24.6      &   63.9    &  19.9 & 97.6\%   &  \textbf{25.6}	 & 25.0	 & 8.1& 22.9\%
      \\

                                           & Gist-COCO   
                                           &\textbf{34.9} &\textbf{28.9} &\textbf{69.6} &\textbf{22.6}& 99.1\% 

                                           &24.7	&\textbf{25.3}	&\textbf{8.8}
                                           
                                           & 35.9\%
\\\cline{2-11}

                                           & Full Prompt                          &   43.3      &  33.5   &  75.1     &  25.4  & -  &36.0	 &34.4	 &12.5 & -
         \\ \hline

\multirow{4}{*}{Llama2-7b}          & No Prompt                                                   &  26.0       & 24.0        &    67.8      &     20.9 & - &21.3 &20.8 &6.2	& -	                \\
                                        & SEGENC~\shortcite{vig2021exploring}                                  &  29.9       &  29.2      &   70.6   &  21.8 & 97.6\%  &\textbf{26.1}	&24.7	&\textbf{8.6} & 22.9\%
         \\

                                          & Gist-COCO   
                                          &\textbf{35.9} &\textbf{30.8} &\textbf{71.9} &\textbf{24.4} & 99.1\% 
                                           &22.7	&\textbf{24.9} &	8.4

                                          & 35.9\%

                                         \\\cline{2-11}
                                      
                                        & Full Prompt                         &  45.2       & 35.0       &   75.4     &  27.9 & - &35.5	&32.8	&12.3& -
     \\ \hline

\multirow{4}{*}{Llama-13b}          & No Prompt                                                   & 27.6         &  27.2       &  72.9        &   22.0 & - &22.3 &  18.7  &4.0   & -           \\
                                
                                        & SEGENC~\shortcite{vig2021exploring}                                  &  31.3      & 29.2      & 70.5      &   22.7 & 97.6\% &\textbf{27.7}	&\textbf{26.2}	&9.5 & 22.9\%
          \\

                                        & Gist-COCO  
                                         &\textbf{36.9} &\textbf{30.8} & \textbf{74.5}  &\textbf{24.4}& 
                                         99.1\% 
                                          &24.8	 &26.0	 &\textbf{9.6}

                                        & 35.9\%
     \\\cline{2-11}

                                         & Full Prompt                           &   45.7    &  36.0   &  77.6      &  29.3    & -  & 37.6	& 38.0	& 14.4& -
   \\ \hline

\end{tabular}
}
\caption{Effectiveness of Prompt Compression on Decoder-based Language Models. }
\label{tab:diff_basemodel}
\end{table*}
\subsection{Overall Performance}
The experiments show the effectiveness of Gist-COCO in the tasks of passage compression and instruction compression, utilizing both encoder-decoder-based language models and decoder-based language models for evaluation. 

The representation-based prompt compression performance is shown in Table~\ref{tab:overall}. In our experiments, we implement the Gist and AutoCompressor models by training a linear layer to adapt the compressed representations to the FlanT5 model. When compared to the fully finetuned compression model, Gist (Ours), they demonstrate comparatively less effectiveness in assisting FlanT5 to comprehend the knowledge and user intent conveyed through the prompts. This suggests that representation-based compression models still require finetuning to tailor them to different language models, limiting the generalization ability of these baseline models. 

The evaluation results show that Gist-COCO outperforms all compression baseline models, demonstrating its ability to learn more tailored gist representations for prompt compression. Notably, Gist-COCO achieves more than a 20\% improvement on the passage compression task, showing its effectiveness in distilling some necessary information from the raw prompts to the gist representations. Different from the baseline models, such as Gist (Ours), Gist-COCO freezes the parameters of language models and only finetunes an additional encoder model specifically for prompt compression, which helps to preserve the capabilities of vanilla language models. It breaks the limitation of compression generalization by directly using the decoder module of vanilla language models to verbalize the gist representations into gist prompts for aiding different LLMs. 

We then extend the evaluation of Gist-COCO's compression efficacy to decoder-based language models by using gist prompts (Eq.~\ref{eq:verbalization}) to replace raw prompts. The evaluation results are shown in Table~\ref{tab:diff_basemodel}. Overall, Gist-COCO enhances the generation accuracy of Llama-7b/13b by furnishing compressed prompts, demonstrating their ability to extract essential information from raw prompts. In comparison to the query-focused passage compression model, SEGENC, Gist-COCO achieves competitive or even superior performance in both passage and instruction compression tasks. This highlights the capacity of leveraging the language model itself for prompt compression and selecting informative contents in an unsupervised manner.

\begin{table}
\begin{center}
\small
\resizebox{\linewidth}{!}{
\begin{tabular}{l|c|c|c|c}
\hline
             \textbf{Setting}  & \textbf{\#Token}  &\textbf{PopQA}  &\textbf{KILT} &\textbf{Alpaca+} \\ \cline{1-5}
\multirow{3}{*}{Unified}  & 5  &24.3 &30.3 &24.7       \\
 & 10  &26.6 &32.1&24.6 \\
 & 20  &\textbf{30.9} &\textbf{35.8}&\textbf{26.5} \\
\hline

\multirow{5}{*}{\shortstack[l]{ Gist-COCO \\ (Disentangled)}}  & 1   &16.8 & 24.2&19.2  \\ 

& 5   &27.4 & 33.1&24.7  \\ 
                                        
& 10    &32.0 &36.2&26.3  \\
& 15    &34.5 & 37.2&\textbf{26.8}  \\
& 20    &\textbf{35.8} &\textbf{38.0}&26.7 \\ 
\hline
\end{tabular}}
\caption{Ablation Studies. We employ varying numbers of gist tokens to encode prompts as hidden states and feed them to FlanT5-large for evaluating the compression effectiveness.}
\label{tab:ablation}
\end{center}
\end{table}

\subsection{Ablation Studies}
This experiment conducts ablation studies to demonstrate the effectiveness of Gist-COCO with varying numbers of gist tokens and explores the impact of employing unified gist tokens. More ablation studies are shown in Appendix~\ref{app:abalation_study}.


As shown in Table~\ref{tab:ablation}, we conduct the Unified and Gist-COCO (Disentangled) settings to train the model to compress the prompts into gist tokens, separately. In the unified setting, we utilize all gist tokens to compress both passages and instructions. Our Gist-COCO model uses disentangled gist tokens that are allocated in equal numbers for compressing passages and instructions. For example, the number of gist tokens in the decomposition setting is 5 signifies that we use 5 gist tokens to compress passages and another 5 gist tokens to compress instructions.

The evaluation results show that, disentangling the gist tokens for various compression tasks typically leads to improvements, highlighting the necessity of utilizing distinct gist tokens to represent various tasks. As the number of gist tokens increases, the compression performance strengthens accordingly. This indicates that additional gist tokens can capture and convey more information from prompts and inputs, thereby enhancing the language model generation process. However, it's noteworthy that the performance improvement tends to plateau after reaching a gist token count of 10. Consequently, we opt for 10 as the optimal gist token count for compressing both passages and instructions.

\begin{figure}
  \centering
  \includegraphics[width=0.3\textwidth]{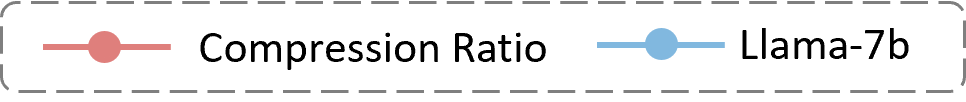}\label{fig:embed:mlm}\\
  \subfigure[PopQA.]
  {\includegraphics[width=0.23\textwidth]{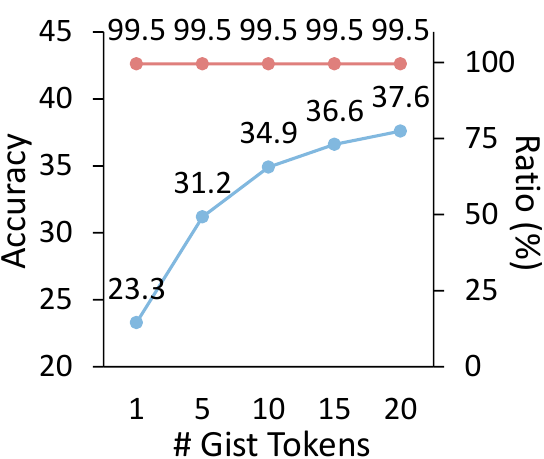}\label{fig:embed:codet5}}
  \subfigure[NQ.]{\includegraphics[width=0.23\textwidth]{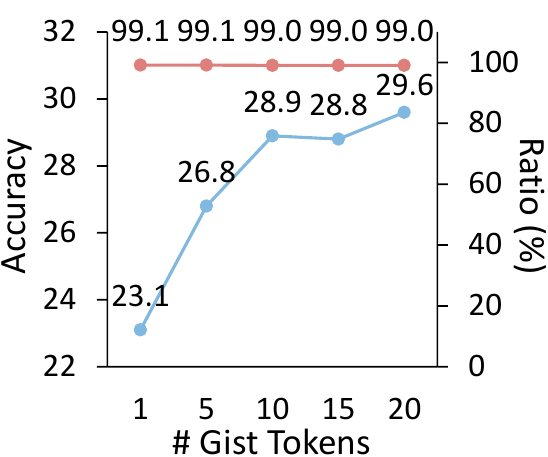}\label{fig:embed:CL}}
  \subfigure[Alpaca+: Seen.]{\includegraphics[width=0.23\textwidth]{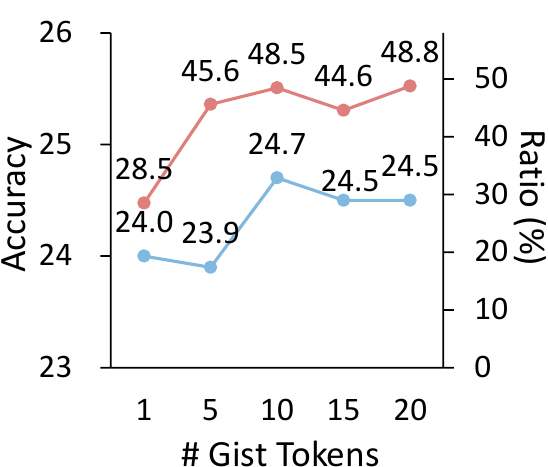}\label{fig:embed:mlm}}
\subfigure[Alpaca+: Human.]{\includegraphics[width=0.23\textwidth]{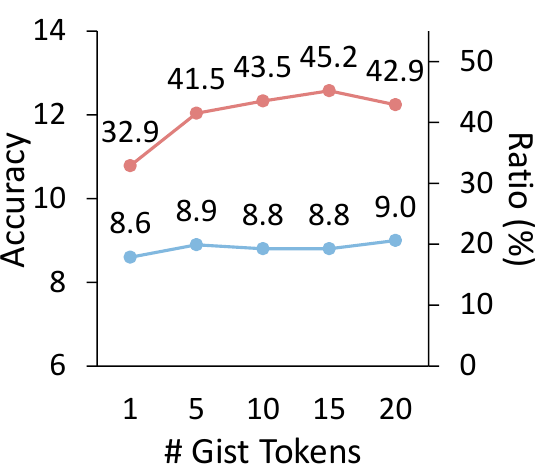}\label{fig:embed:mlm}}
  
  \caption{Effectiveness of Gist Verbalization Results. We use different numbers of compression tokens.}
  \label{fig:gist_visualization}
\end{figure}
Then we show the effectiveness of the verbalization outputs produced by Gist-COCO, as depicted in Figure~\ref{fig:gist_visualization}, utilizing the Llama-7b model. Evaluation results indicate that the verbalized outputs from Gist-COCO consistently enhance the performance of Llama-7b as the number of gist tokens increases. Conversely, performance remains almost unchanged across instruction compression tasks. This illustrates that passages typically encompass more compressible information, while 10 gist tokens are adequate for instruction compression. Moreover, the compression ratio remains stable across different numbers of gist tokens, indicating that prompts are typically treated as short prefixes for language models, and certain tokens play a more crucial role in aiding language models.

\begin{figure}[t]
    \centering
    \subfigure[Passage Compression.] { \label{fig:embed:compression} 
    \includegraphics[width=0.48\linewidth]{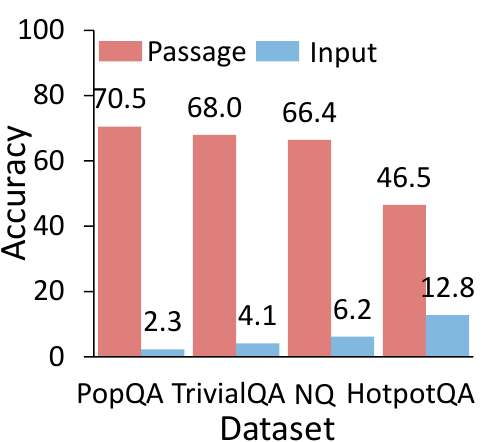}}
    \subfigure[Instruction Compression.] { \label{fig:tasks} 
    \includegraphics[width=0.48\linewidth]{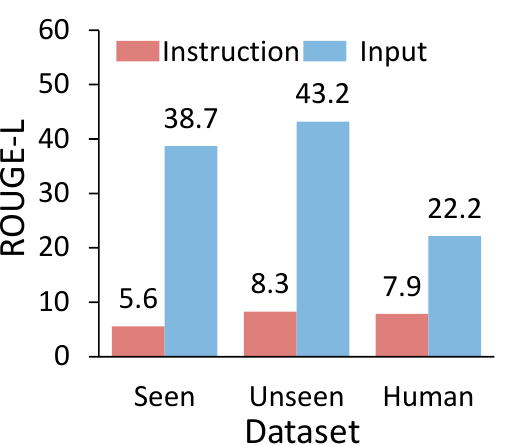}}
    \caption{Text Similarity between the Gist Verbalization Results with Inputs and Prompts.}
    \label{fig:similarity}
\end{figure}

\begin{figure}[t]
\centering
 \includegraphics[width=\linewidth]{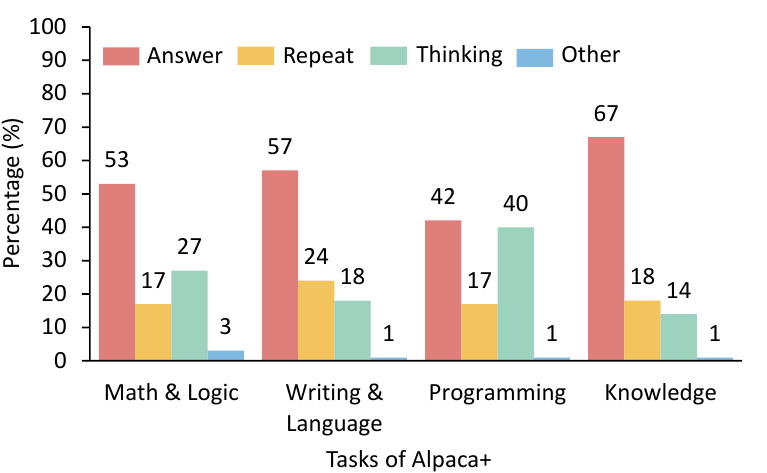}\label{fig:embed:CL}
\caption{Distribution of Categorizations of Gist Verbalization Results. We categorize Alpaca+ tasks into distinct groups and present the categorization outcomes of verbalization results across various tasks.}
\label{fig:category}
\end{figure}

\subsection{Characteristics of Learned Gist Representations}
In this experiment, by verbalizing these gist representations into gist prompts, we further analyze the knowledge learned by gist tokens.

As shown in Figure~\ref{fig:similarity}, we first evaluate the text similarity between the gist prompts and both inputs and prompts. Regarding the passage compression tasks, the gist prompts exhibit a notably high resemblance to the passages rather than the inputs. This observation underscores that the primary objective of passage compression is to extract essential knowledge from the passage to facilitate question answering. In contrast, for the tasks in Alpaca+, the gist prompts demonstrate much higher similarity to the inputs. This suggests that our Gist-COCO model engages in a more profound analysis of the queries using the provided instructions. 

Then we explore the roles of gist prompts across various tasks in Figure~\ref{fig:category}. We firstly employ GPT-3.5 to categorize the data within the Alpaca+ dataset into four distinct groups: Match \& Logic, Writing \& Language, Programming, and Knowledge. Detailed categorization statistical information is shown in Appendix~\ref{app:Classification}. Subsequently, we randomly select 100 instances from each task group and assign labels to the sampled data using GPT-3.5. These labels include Answer, Repeat, Thinking, and Other. The ``Answer'' label denotes that the gist prompts provide potential answers to the input. The ``Thinking'' label signifies that the gist prompts serve as a form of thought process. Meanwhile, the ``Repeat'' label indicates that the gist prompts reiterate the content of queries or instructions.

The evaluation results indicate that directly generating answers is the predominant behavior across different tasks. It demonstrates that compression models usually serve as a form of information preprocessing to give the answer-like results to aid language models. Across all tasks, Gist-COCO tends to repeat prompts or inputs more frequently in the Writing \& Language tasks, underscoring the significance of user intent in the task. Moreover, there is a preference for generating a chain of thought to aid Match \& Logic and Programming tasks, highlighting the critical role of the thought process in dealing with these tasks~\cite{wei2022chain,li2023structured,huang2023codecot}.

\begin{table*}[t]
\label{tab:case_study}
\small
\begin{tabular}{p{0.17\linewidth}|p{0.79\linewidth}} 
\hline
\multicolumn{2}{l}{\textbf{\textit{Passage Compression}}}
\tabularnewline 
\hline
\makecell[l]{PopQA} & \makecell[l]{\textbf{Passage:} Page 3 (film)
Page 3 is a 2005 Indian drama film directed by \textcolor[rgb]{0.7,0.3,0.3}{\textbf{Madhur Bhandarkar}} and \\produced by Bobby Pushkarna and Kavita Pushkarna about the Page 3 culture and media in the city \\of Mumbai. It stars Konkona Sen Sharma, Atul Kulkarni, Sandhya Mridul, Tara Sharma, Anju \\Mahendru, and Boman Irani. The film won three National Film Awards ...\\
\textbf{Input:} Who was the director of Page 3?\\
\textbf{Compression:} \textcolor[rgb]{0.7,0.3,0.3}{\textbf{Madhur Bhandarkar}}}
\tabularnewline 
\hline
\makecell[l]{TriviaQA} & \makecell[l]{\textbf{Passage:}...The screenplay by Robert E. Sherwood and Joan Harrison, and adaptation by \\Philip MacDonald and Michael Hogan, were based on the 1938 novel of the same name by\\ \textcolor[rgb]{0.7,0.3,0.3}{\textbf{Daphne du Maurier}}. The film stars Laurence Olivier as the brooding...\\
\textbf{Input:} The Alfred Hitchcock films Rebecca and The Birds were based on novels by which author?\\
\textbf{Compression:} \textcolor[rgb]{0.7,0.3,0.3}{\textbf{Daphne du Maurier}}}
\tabularnewline 
\hline
\multicolumn{2}{l}{\textbf{\textit{Instruction Compression}}}\\
\hline
\makecell[l]{Knowledge} & \makecell[l]{\textbf{Instruction:} Classify this sentence into one of the topics: education, politics, \textcolor[rgb]{0.7,0.3,0.3}{\textbf{technology}}, sports \\
\textbf{Input:} Apple's new Iphone was released today.\\
\textbf{Compression:} \textcolor[rgb]{0.7,0.3,0.3}{\textbf{technology}}}
\tabularnewline 
\hline
\makecell[l]{Match \& Logic} & \makecell[l]{\textbf{Instruction:} What is the best way to get from point a to point b? explain why you chose that method.\\
\textbf{Input:} Point A: (0, 0) and Point B: (10, 10)\\
\textbf{Compression:} The first step is to get to the point where you want to go.}
\tabularnewline 
\hline
\makecell[l]{Programming} & \makecell[l]{\textbf{Instruction:} You are given a programming problem and its implementation. Analyze the problem \\and implementation and explain the algorithm and approach used to solve the problem. \\
\textbf{Input:} Table: Person \textbackslash n| Column Name | Type |\textbackslash n | personId  | int  |\textbackslash n | lastName  | varchar |\textbackslash\\ n | firstName | varchar |...\\
\textbf{Compression:} SELECT T1.name FROM Person AS T1 JOIN Address AS T2 ON T1.name = \\T2.name JOIN Person AS T}
\tabularnewline 
\hline
\makecell[l]{Writing \& Language } & \makecell[l]{\textbf{Instruction:} The topic of YouTube post has been described and based on the information, you need\\ to write a hook for starting the post. A catchy hook will keep readers interested so they keep reading.  \\
\textbf{Input:} \textcolor[rgb]{0.7,0.3,0.3}{\textbf{A video showing how to make a tasty}} cup of \textcolor[rgb]{0.7,0.3,0.3}{\textbf{coffee.}}\\
\textbf{Compression:}  \textcolor[rgb]{0.7,0.3,0.3}{\textbf{A video showing how to make a tasty coffee.}}}
\tabularnewline 
\hline

\end{tabular}
\caption{Case Studies. The matched text phrases are \textcolor[rgb]{0.7,0.3,0.3}{\textbf{highlighted}}.}
\label{tab:case_study}
\end{table*}

\subsection{Case Studies}
Finally, we show several cases in Table~\ref{tab:case_study} to analyze the gist prompts of Gist-COCO.

In the first two cases, the gist prompts like ``Madhur Bhandarkar'' and ``Daphne du Maurier'' indicate that the extracted segments from the passage can directly answer the question. It demonstrates the compression module's tendency to directly generate answers for simpler questions, highlighting its preprocessing capabilities. For the third and fourth cases, involving mathematical and programming tasks, strategic planning and critical thinking are necessary. Gist-COCO shows its effectiveness in generating preliminary thoughts or code snippets as prompts to assist language models in comprehending and solving such problems. It confirms that the chain-of-thought and program thought indeed have the ability to improve the model's effectiveness on these tasks. The final case illustrates a writing and language task, where the results indicate Gist-COCO's inclination to replicate the input, suggesting the continued challenge in verbalizing and analyzing such instructions.



\section{Conclusion}
This paper introduces Gist-COCO, a prompt compression approach utilizing gist conditioned decoding. Our experiments demonstrate that Gist-COCO surpasses existing compression models across various prompt compression tasks and extends its effectiveness to different language models. Further analyses provide some opportunities to understand the prompt behaviors in language models, facilitating a deeper understanding of their functionality.

\section*{Limitations}
Although Gist-COCO has demonstrated considerable success in compression prompts, it encounters inherent limitations. Existing prompt compression is still difficult to achieve the same results as the original prompt with a high compression ratio, and there are still different degrees of information loss in the prompt compression process. To mitigate this, Gist-COCO attempts to increase the number of gist tokens, but the improvement is limited. 

Besides, there are some instructions that are hard to compress, making Gist-COCO repeat the contents in the inputs. In this case, it is still challenging to interpret which contents can really assist the language models to follow the given instruction.


\bibliography{acl_main}
\clearpage
\newpage
\appendix
\section{Appendix}
\subsection{License}
We show the licenses of the datasets that we use. PopQA, MS MARCO and KILT use MIT license. Alpaca+ and NVI2 use Apache license. All of these licenses and agreements allow their data for academic use.

\subsection{Additional Ablation Studies on Gist-COCO}\label{app:abalation_study}
\begin{table*}[t]
\small
\centering
\begin{tabular}{l|l|c|c|ccc|ccc}
\hline
\multirow{3}{*}{\textbf{LLM}} & \multicolumn{1}{l|}{\multirow{3}{*}{\textbf{Setting}}} &  \multicolumn{1}{l|}{\multirow{3}{*}{\textbf{\#Token}}}  & \multicolumn{4}{c|}{\textbf{Passage Compression}} & \multicolumn{3}{c}{\textbf{Instruction Compression}} \\  \cline{4-10} &   &   & \multirow{2}{*}{\textbf{PopQA}}  & \multicolumn{3}{c}{\textbf{KILT}}  & \multicolumn{3}{c}{\textbf{Alpaca+}} \\\cline{5-10}  

                                        &     &                        &            &  NQ        & TrivialQA        & HotpotQA   &  Seen        & Unseen   &Human                   \\ \hline

\multirow{8}{*}{FlanT5-large}    & \multirow{3}{*}{Unified}  &5  &24.3 &21.2 &47.6 &18.3 &24.0 &28.1  & 14.0        \\
& &10  &26.6 &22.8 &50.7 &19.1 &23.6 &28.4 & 13.3       \\
 & &20  &30.9 &26.0 &56.6 &20.8 &24.3 &29.6 & 14.1      \\ \cline{2-10}

                                       &\multirow{5}{*}{Disentangled} &1   &16.8&15.1&36.9& 16.7
 
                                       &20.4	&20.2	&10.4
\\ 

& &5   &27.4 &23.7 &52.4 &19.3 

&24.8	&27.4	&13.6
\\ 
                                        
                                      &  &10   &32.0 &27.0 &57.3 &20.6 
    
                                      &25.7	&30.1	&14.0
\\ 
&  &15   &34.5 &27.8 &58.8 &21.3 

 &25.4	 &\textbf{31.3}	 &14.9
\\ 

 &   &20   &\textbf{35.8} &\textbf{28.6} &\textbf{59.9}&\textbf{21.8} 
& \textbf{25.6}	&30.6	&\textbf{15.2}

   \\ \hline

\multirow{8}{*}{Llama-7b}     & \multirow{3}{*}{Unified} &5      &29.2 &25.9 &67.4 &21.5

 &24.5	 &24.9	 &7.7
                              \\
& &10       &31.0&25.4&66.2&21.1&24.0  & 24.9&7.3                        \\
& &20       &33.5&27.8&68.6&22.0&21.7  &25.2 &7.9    
\\\cline{2-10}

 &\multirow{5}{*}{Disentangled} &1                            &23.3&23.1&62.8&19.4
 &24.0	&23.8	&8.6
    \\

                                         &   &5                            &31.2 &26.8 &68.5 &22.2 
                                         &23.9	&25.2	&8.9

    \\

                                          &   &10                            &34.9&28.9 &69.6 &22.6
                                          &\textbf{24.7}	&25.3	&8.8

                                            \\

                                           &  &15                            &36.6 &28.8&70.1&22.8
                                           
                                           &24.5	&25.4	&8.8
 \\

                                            & &20                            &\textbf{37.6} &\textbf{29.6}&\textbf{70.3}&\textbf{22.8}
                                            &24.5	&\textbf{25.4}	&\textbf{9.0}

                                            \\\hline
\multirow{8}{*}{Llama2-7b}                   &\multirow{3}{*}{Unified} &5   &30.6 &28.5 &70.2 &23.5 
&22.3	&23.8	&7.6
                          
                                         \\
 & &10  &32.8&28.7&69.8&22.9&21.9&24.2 &7.6  \\ 
 &  &20  &34.4&29.7&71.2&24.1&21.6&24.5 &6.9  \\ \cline{2-10}
 
 & \multirow{5}{*}{Disentangled} &1                  &25.3&26.4&67.8&22.0 
 &20.7	&21.7	&7.5

 \\
                     &  &5                           &32.7 &29.4 &71.2 &23.9
                     &21.6	&24.2	&7.4

\\
                      &   &10                          &35.9&30.8 &71.9 &24.4 
                     & \textbf{22.7}	&\textbf{24.9}	&8.4\\
                     &     &15                        &37.6&31.4&72.3&\textbf{24.8}
                      &22.7	 &24.7	 &7.8

                \\

                                           &  &20                            & \textbf{38.4}&\textbf{31.4}&\textbf{72.7}&24.7
                                           &22.0	&25.1	&\textbf{8.8}

                         \\\hline

\multirow{8}{*}{Llama-13b}    &\multirow{3}{*}{Unified} &5    &32.4 &28.0&73.7 &23.5 &24.6	&25.7	&8.6
                                       \\

 & &10       &34.0 &28.4&73.0 &23.7 &24.2  &25.4 &8.0                       \\
 & &20                           &35.8&29.8 &74.4 &24.1&23.4&\textbf{26.1} &7.6   \\\cline{2-10}

  &\multirow{5}{*}{Disentangled} &1   &27.0&26.0&71.7&23.2
  &24.0	&23.7	&9.0

             \\
                    & &5   & 34.0     &29.3  &74.0 & 24.0  
                    &24.7	&25.6	&9.0

             \\
                   &   &10   &36.9   &30.8   &74.5  &24.4 
                   
                   & \textbf{24.8}	 &26.0	 &9.6
\\

                  &    &15  &38.4 &30.8 &74.3 &\textbf{24.6} 

                  &24.6	&25.8	&\textbf{9.7}

                  \\
                    &   &20  &\textbf{39.3} &\textbf{31.0} &\textbf{74.5} &24.4 
                    &24.4	&25.7	&9.6

                  \\
                     
                     \hline

\end{tabular}
\caption{Additional Ablation Studies. For FlanT5-large, we employ the embedding-based compression modeling method, as well as verbalize the gist representations as prompts for other models.}
\label{tab:more_abalation}
\end{table*}
We conduct additional ablation studies to delicately explore the compression effectiveness on different models with different gist modeling methods.

As shown in Table~\ref{tab:more_abalation}, the disentangled gist modeling method is more effective than the unified gist modeling method, when generalizing the compression capabilities of Gist-COCO to different LLMs. With an increase in the number of gist tokens, there is an enhancement in gist verbalization performance. However, once the number of gist tokens surpasses 10, the rate of improvement slows down, impacting performances on certain tasks.

\begin{table}
\begin{center}
\small
\begin{tabular}{l|l}
\hline
\textbf{Type Name}    &\textbf{Total}    \\\hline
           Math \& Logic Problems             & 504        \\Writing \& Language  Problems& 956 \\

Programming Problems & 273\\

Knowledge & 519 \\
\hline
\end{tabular}

\caption{Data Statistics of Different Classifications of Alpaca+ Data.}
\label{tab:gist_type}
\end{center}
\end{table}
\subsection{Data Classification of Alpaca+ Data}\label{app:Classification}
We employ ChatGPT-3.5 to categorize the data within the Alpaca+ dataset into four distinct groups. The data statistics are shown in Table~\ref{tab:gist_type}.

\end{document}